\documentclass[a4paper,twoside,french]{article}
\usepackage{rfia2000}
\usepackage[T1]{fontenc}
\usepackage{babel}
\usepackage{times}

\usepackage{subfigure}

\usepackage{amsmath}

\usepackage[pdftex]{graphicx}
\graphicspath{{./images/}}
\DeclareGraphicsExtensions{.png,.jpg,.pdf,.eps}

\begin{document}
\date{}

\title{\Large\bf Combinaison d'information visuelle, conceptuelle, et contextuelle pour la construction automatique de hiérarchies sémantiques adaptées à l'annotation d'images}


\author{\begin{tabular}[t]{c@{\extracolsep{8em}}c}
Hichem Bannour & C\'eline Hudelot \\
\end{tabular}
{} \\
 \\
Laboratoire de Mathématiques Appliquées aux Systèmes (MAS)\\
École Centrale Paris
{} \\
 \\
Grande Voie des Vignes\\
92295 Châtenay-Malabry, France\\
\{Hichem.bannour, Celine.hudelot\}@ecp.fr
}
\maketitle
\thispagestyle{empty}
\subsection*{R\'esum\'e}
{\em
Ce papier propose une nouvelle méthode pour la construction automatique de hiérarchies sémantiques adaptées à la classification et à l'annotation d'images. La construction de la hiérarchie est basée sur une nouvelle mesure de similarité sémantique qui intègre plusieurs sources d'informations: visuelle, conceptuelle et contextuelle que nous définissons dans ce papier. L'objectif est de fournir une mesure qui est plus proche de la sémantique des images. Nous proposons ensuite des règles, basées sur cette mesure, pour la construction de la hiérarchie finale qui encode explicitement les relations hiérarchiques entre les différents concepts. La hiérarchie construite est ensuite utilisée dans un cadre de classification sémantique hiérarchique d'images en concepts visuels. Nos expériences et résultats montrent que la hiérarchie construite permet d'améliorer les résultats de la classification.
}
\subsection*{Mots Clef}
Construction de hiérarchies sémantiques, sémantique d'images, annotation d'images, mesures de similarité sémantiques, classification hiérarchique d'images.

\subsection*{Abstract}
{\em
This paper proposes a new methodology to automatically build semantic hierarchies suitable for image annotation and classification. The building of the hierarchy is based on a new measure of semantic similarity. The proposed measure incorporates several sources of information: visual, conceptual and contextual as we defined in this paper. The aim is to provide a measure that best represents image semantics. We then propose rules based on this measure, for the building of the final hierarchy, and which explicitly encode hierarchical relationships between different concepts. Therefore, the built hierarchy is used in a semantic hierarchical classification framework for image annotation. Our experiments and results show that the hierarchy built improves classification results.
}
\subsection*{Keywords}
Semantic hierarchies building, image semantics, image annotation, semantic relatedness measure, hierarchical image classification.

\section{Introduction}

Avec l'explosion des données images, il devient essentiel de fournir une annotation sémantique de haut niveau à ces images pour satisfaire les attentes des utilisateurs dans un contexte de recherche d'information. Des outils efficaces doivent donc être mis en place pour permettre une description sémantique précise des images. Depuis les dix dernières années, plusieurs approches d'annotation automatique d'images ont donc été proposées \cite{Barnard03, Lavrenko03,Fan08IP,Bannour09CORIA,Bannour10} pour essayer de réduire le problème bien connu du \emph{fossé sémantique} \cite{Smeulders00}. Cependant, dans la plupart de ces approches, la sémantique est souvent limitée à sa manifestation perceptuelle, i.e. au travers de l'apprentissage d'une fonction de correspondance associant les caractéristiques de bas niveau à des concepts visuels de plus haut niveau sémantique \cite{Barnard03,Lavrenko03}. Cependant, malgré une efficacité relative concernant la description du contenu visuel d'une image, ces approches sont incapables de décrire la sémantique d'une image comme le ferait un annotateur humain. Elles sont également confrontées au problème du passage à l'échelle \cite{Liu07}. En effet, les performances de ces approches varient considérablement en fonction du nombre de concepts et de la nature des données ciblées \cite {Hauptmann07}. Cette variabilité peut être expliquée d'une part par la large variabilité visuelle intra-concept, et d'autre part par une grande similarité visuelle inter-concept, qui conduisent souvent à des annotations imparfaites.

Récemment, plusieurs travaux se sont intéressés à l'utilisation de hiérarchies sémantiques pour surmonter ces problèmes \cite{Tousch11,Bannour11,BannourMMM12}. En effet, l'utilisation de connaissances explicites, telles que les hiérarchies sémantiques, peut améliorer l'annotation en fournissant un cadre formel qui permet d'argumenter sur la cohérence des informations extraites des images. En particulier, les hiérarchies sémantiques se sont avérées être très utiles pour réduire le fossé sémantique \cite{Deng10}. Trois types de hiérarchies pour l'annotation et la classification d'images ont été récemment explorées : 1) les hiérarchies basées sur des connaissances textuelles (nous ferons référence à ce type de connaissances par information conceptuelle dans le reste du papier) \footnote{Exemple d'information textuelle utilisée pour la construction des hiérarchies: les tags, contexte environnant, WordNet, Wikipedia, etc.} \cite{Marszalek07,Wei07,Deng09}, 2) les hiérarchies basées sur des informations visuelles (ou perceptuelles), i.e. caractéristiques de bas niveau de l'image \cite{Sivic08,Bart08,Yao09}, 3) les hiérarchies que nous nommerons sémantiques basées à la fois sur des informations textuelles et visuelles \cite{LI10,Fan07,Wu08}. Les deux premières catégories d'approches ont montré un succès limité dans leur usage. En effet, d'un côté l'information conceptuelle seule n'est pas toujours en phase avec la sémantique de l'image, et est alors insuffisante pour construire une hiérarchie adéquate pour l'annotation d'images \cite{Wu08}. De l'autre coté, l'information perceptuelle ne suffit pas non plus à elle seule pour la construction d'une hiérarchie sémantique adéquate (voir le travail de \cite{Sivic08}). En effet, il est difficile d'interpréter ces hiérarchies dans des niveaux d'abstraction plus élevés. Ainsi, la combinaison de ces deux sources d'information semble donc obligatoire pour construire des hiérarchies sémantiques adaptées à l'annotation d'images.

La suite de ce papier est organisée comme suit: dans la section 2 nous présentons les travaux connexes. La section 3 présente la mesure sémantique proposée dans un premier temps, puis les règles utilisées pour la construction de la hiérarchie sémantique. Les résultats expérimentaux sont présentés dans la section 4. La section 5 présente nos conclusions et perspectives.

\section{État de l'art}

Plusieurs méthodes \cite{LI10,Fan07,Marszalek07,Wei07,Sivic08,Bart08} ont été proposées pour la construction de hiérarchies de concepts dédiées à l'annotation d'images. Dans cette section nous présenterons ces différentes méthodes en suivant l'ordre proposé dans l'introduction.

Marszalek \& al. \cite{Marszalek07} ont proposé de  construire une hiérarchie par l'extraction du graphe pertinent dans WordNet reliant l'ensemble des concepts entre eux. La structure de cette hiérarchie est ensuite utilisée pour construire un ensemble de classifieurs hiérarchiques. Deng \& al. \cite{Deng09} ont proposé \emph{ImageNet}, une ontologie à grande échelle pour les images qui repose sur la structure de WordNet, et qui vise à peupler les 80 000 synsets de WordNet avec une moyenne de 500 à 1000 images sélectionnées manuellement. L'ontologie LSCOM \cite{LSCOM06} vise à concevoir une taxonomie avec une couverture de près de 1 000 concepts pour la recherche de vidéo dans les bases de journaux télévisés. Une méthode pour la construction d'un espace sémantique enrichi par les ontologies est proposée dans \cite{Wei07}. Bien que ces hiérarchies soient utiles pour fournir une structuration compréhensible des concepts, elles ignorent l'information visuelle qui est une partie importante du contenu des images.

D'autres travaux se sont donc basés sur l'information visuelle \cite{Sivic08,Bart08,Yao09}. Une plateforme (I2T) dédiée à la génération automatique de descriptions textuelles pour les images et les vidéos est proposée dans \cite{Yao09}. I2T est basée principalement sur un graphe AND-OR pour la représentation des connaissances visuelles. Sivic \& al. \cite{Sivic08} ont proposé de regrouper les objets dans une hiérarchie visuelle en fonction de leurs similarités visuelles. Le regroupement est obtenu en adaptant, pour le domaine de l'image, le modèle d'Allocation Dirichlet Latente hiérarchique (hLDA) \cite{Blei04}. Bart \& al. \cite{Bart08} ont proposé une méthode bayésienne pour organiser une collection d'images dans une arborescence en forme d'arbre hiérarchique. Dans \cite{Griffin08}, une méthode pour construire automatiquement une taxonomie pour la classification d'images est proposée. Les auteurs suggèrent d'utiliser cette taxonomie afin d'augmenter la rapidité de la classification au lieu d'utiliser un classifieur multi-classe sur toutes les catégories. Une des principales limitations de ces hiérarchies visuelles est qu'elles sont difficiles à interpréter. Ainsi, une hiérarchie sémantique compréhensible et adequate pour l'annotation d'images devrait tenir compte à la fois de l'information conceptuelle et de l'information visuelle lors du processus du construction.

Parmi les approches pour la construction de hiérarchies sémantiques, Li \& al. \cite{LI10} ont présenté une méthode basée à la fois sur des informations visuelles et textuelles (les étiquettes associées aux images) pour construire automatiquement une hiérarchie, appelée "semantivisual", selon le modèle hLDA. Une troisième source d'information que nous nommerons information contextuelle est aussi utilisée pour la construction de telles hierarchies. Nous discutons plus précisément de cette information dans le paragraphe suivant. Fan \& al. \cite {Fan09} ont proposé un algorithme qui intègre la similarité visuelle et la similarité contextuelle entre les concepts.  Ces similarités sont utilisées pour la construction d'un réseau de concepts utilisé pour la désambiguïsation des mots. Une méthode pour la construction de hiérarchies basées sur la similarité contextuelle et visuelle est proposée dans \cite{Fan07}. La "distance de Flickr" est proposée dans \cite{Wu08}. Elle représente une nouvelle mesure de similarité entre les concepts dans le domaine visuel. Un réseau de concepts visuels (VCNet) basé sur cette distance est également proposé dans \cite{Wu08}. Ces hiérarchies sémantiques ont un potentiel intéressant pour améliorer l'annotation d'images.

\textbf{Discussion}
\label{sec:Disc}

Comme nous venons de le voir, plusieurs approches de construction de hierarchies se basent sur WordNet \cite{Marszalek07,Deng09}. Toutefois, WordNet n'est pas très approprié à la modélisation de la sémantique des images. En effet, l'organisation des concepts dans WordNet suit une structure psycholinguistique, qui peut être utile pour raisonner sur les concepts et comprendre leur signification, mais elle est limitée et inefficace pour raisonner sur le contexte de l'image ou sur son contenu. En effet, les distances entre les concepts similaires dans WordNet ne reflètent pas nécessairement la proximité des concepts dans un cadre d'annotation d'images. Par exemple, selon la distance du plus court chemin dans WordNet, la distance entre les concepts "Requin" et "Baleine" est de 11 (n\oe uds), et entre "Humain" et "Baleine" est de 7. Cela signifie que le concept "Baleine" est plus proche (similaire) de "Humain" que de "Requin". Ceci est tout à fait cohérent d'un point de vue biologique, parce que "Baleine" et "Humain" sont des mammifères tandis que "Requin" ne l'est pas. Cependant, dans le domaine de l'image il est plus intéressant d'avoir une similarité plus élevée entre "Requin" et "Baleine", puisqu'ils vivent dans le même environnement, partagent de nombreuses caractéristiques visuelles, et il est donc plus fréquent qu'on les retrouve conjointement dans une même image ou un même type d'images (ils partagent un même contexte). Donc, une hiérarchie sémantique appropriée devrait représenter cette information ou permettre de la déduire, pour aider à comprendre la sémantique de l'image.

\section{Méthode Proposée}

En se basant sur la discussion précédente, nous définissons les hypothèses suivantes sur lesquelles repose notre approche:\\
\emph{Une hiérarchie sémantique appropriée pour l'annotation d'images doit: 1) modéliser le contexte des images (comme défini dans la section précédente), 2) permettre de regrouper des concepts selon leurs caractéristiques visuelles et textuelles, 3) et refléter la sémantique des images, i.e. l'organisation des concepts dans la hiérarchie et leurs relations sémantiques est fidèle à la sémantique d'images.}

\begin{figure}[!b]
\center
\includegraphics[width=3.2in]{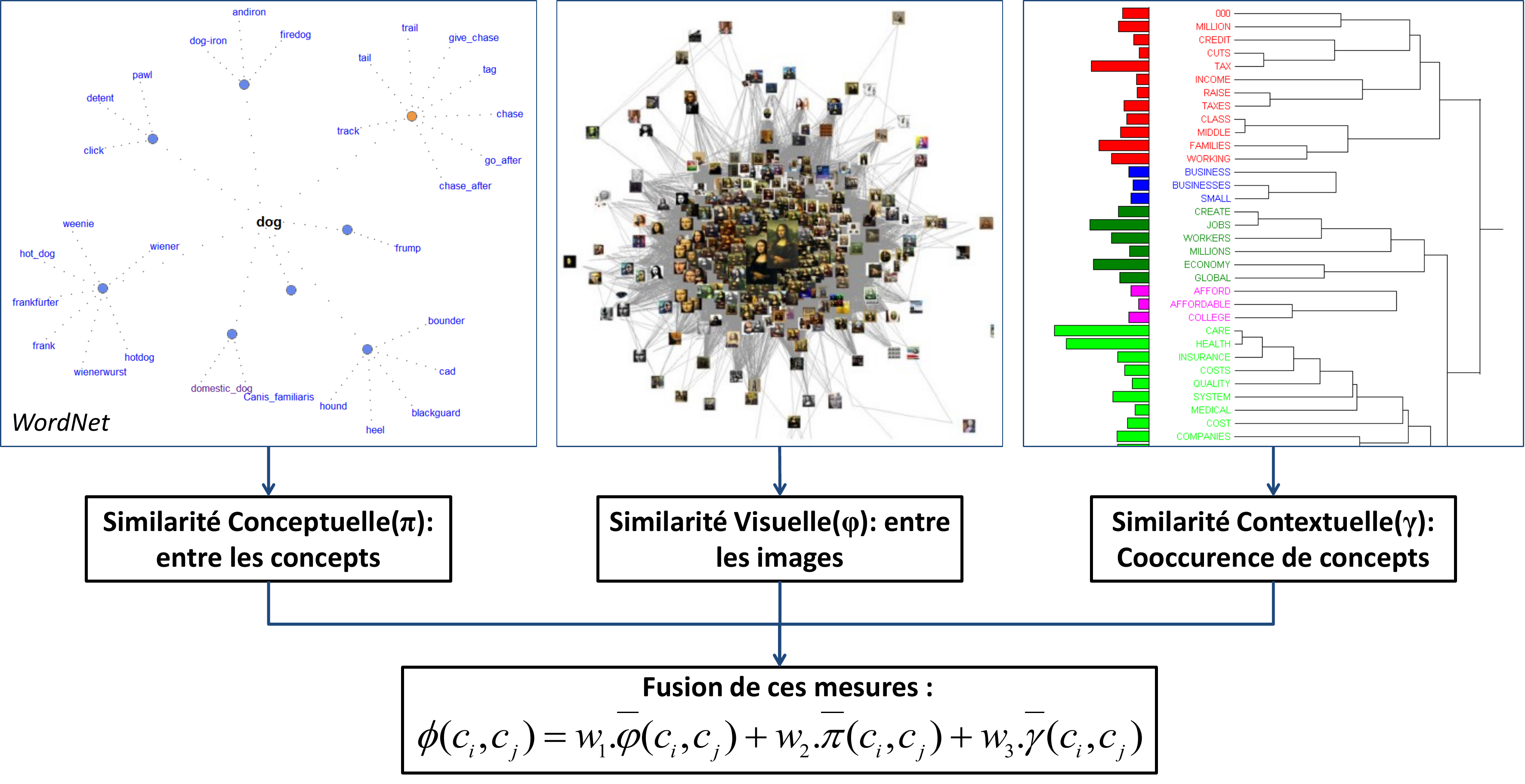}
\caption{Illustration de la mesure proposée basée sur les similarités normalisées: visuelle $\overline{\varphi}$, conceptuelle $\overline{\pi}$ et contextuelle $\overline{\gamma}$ entre concepts.} \label{fig:Proposal}
\end{figure}

Nous proposons dans ce papier une nouvelle méthode pour la construction de hiérarchies sémantiques appropriées à l'annotation d'images. Notre méthode se base sur une nouvelle mesure pour estimer les relations sémantiques entre concepts. Cette mesure intègre les trois sources d'information que nous avons décrites précédemment. Elle est donc basée sur 1) une similarité visuelle qui représente la correspondance visuelle entre les concepts, 2) une similarité conceptuelle qui définit un degré de similarité entre les concepts cibles, basée sur leur définition dans WordNet, et 3) une similarité contextuelle qui mesure la dépendance statistique entre chaque paire de concepts dans un corpus donné (cf. Figure \ref{fig:Proposal}). Ensuite cette mesure est utilisée dans des règles qui permettent de statuer sur la vraisemblance des relations de parenté entre les concepts, et permettent de construire une hiérarchie.

Étant donné un ensemble de couples image/annotation, où chaque annotation décrit un ensemble de concepts associés à l'image, notre approche permet de créer automatiquement une hiérarchie sémantique adaptée à l'annotation d'images. Plus formellement, nous considérons $I=<i_1,i_2,\cdots,i_\mathcal{L}>$ l'ensemble des images de la base considérée, et $C=<c_1,c_2,\cdots,c_\mathcal{N}>$ le vocabulaire d'annotation de ces images, i.e. l'ensemble de concepts associés à ces images. L'approche que nous proposons consiste alors à identifier $\mathcal{M}$ nouveaux concepts qui permettent de relier tous les concepts de $C$ dans une structure hiérarchique représentant au mieux la sémantique d'images.

\subsection{Similarité Visuelle}
\label{sec:Similarité_visuelle}

Soit $x_i^v$ une représentation visuelle quelconque de l'image $i$ (vecteur de caractéristiques visuelles), on apprend pour chaque concept $c_j$ un classifieur qui permet d'associer ce concept à ses caractéristiques visuelles. Pour cela, nous utilisons $\mathcal{N}$ machines à vecteurs de support (SVM) \cite {Vapnik95} binaires (un-contre-tous) avec une fonction de décision $\mathcal{G}(x^v)$:
\begin{equation}\label{Equ:SVM}
    \mathcal{G}(x^v)=\sum_k \alpha_k y_k \mathbf{K}(x_k^v,x^v)+b
\end{equation}
où: $\mathbf{K}(x_i^v,x^v)$ est la valeur d'une fonction noyau pour l'échantillon d'apprentissage $x_i^v$ et l'échantillon de test $x^v$, $y_i \in \{1,-1\}$ est l'étiquette de la classe de $x_i^v$, $\alpha_i$ est le poids appris de l'échantillon d'apprentissage $x_i^v$, et $b$ est un paramètre seuil appris. Il est à noter que les échantillons d'apprentissage $x_i^v$ avec leurs poids $\alpha_i > 0$ forment \emph{les vecteurs de support}.

Après avoir testé différentes fonction noyau sur notre ensemble d'apprentissage, nous avons décidé d'utiliser une fonction noyau à base radiale:
\begin{equation}\label{Equ:Kernel}
    \mathbf{K}(x,y)=exp \Big(\frac{\|x-y\|^2}{\sigma^2}\Big)
\end{equation}

Maintenant, compte tenu de ces $\mathcal{N}$ SVM appris où les représentations visuelles des images sont les entrées et les concepts (classes d'images) sont les sorties, nous voulons définir pour chaque classe de concept un centroïde $\vartheta(c_i)$ qui soit représentatif du concept $c_i$. Les centroïdes définis doivent alors minimiser la somme des carrés à l'intérieur de chaque ensemble $S_i$:
\begin{equation}
\label{Equ1:K-Means}
\underset{S}{\operatorname{argmin}}\sum_{i=1}^{\mathcal{N}}\sum_{x_j^v\in S_i} \|x_j^v-\mu_i\|^2
\end{equation}
où $S_i$ est l'ensemble de \emph{vecteurs de support} de la classe $c_i$, $S=\{S_1,S_2,\cdots,S_\mathcal{N}\}$, et $\mu_i$ est la moyenne des points dans $S_i$.

L'objectif étant d'estimer une distance entre ces classes afin d'évaluer leurs similarités visuelles, nous calculons le centroïde $\vartheta(c_i)$ de chaque concept visuel $c_i$ en utilisant:
\begin{equation}\label{Equ:Moyenne}
    \vartheta(c_i)=\frac{1}{|S_i|}\sum_{x_j \in S_i} x_j^v
\end{equation}

La similarité visuelle entre deux concepts $c_i$ et $c_j$, est alors inversement proportionnelle à la distance entre leurs centroïdes respectifs $\vartheta(c_i)$ et $\vartheta(c_j)$:
\begin{equation}\label{Equ:VisuSim}
   \varphi(c_i,c_j)=\frac{1}{1+d(\vartheta(c_i),\vartheta(c_j))}
\end{equation}
où $d(\vartheta(c_i),\vartheta(c_j))$ est la distance euclidienne entre les deux vecteurs $\vartheta(c_i)$ et $\vartheta(c_j)$ définie dans l'espace des caractéristiques visuelles.

\subsection{Similarité Conceptuelle}

La similarité conceptuelle reflète la relation sémantique entre deux concepts d'un point de vue linguistique et taxonomique. Plusieurs mesures de similarité ont été proposées dans la littérature \cite{Budanitsky06,Resnik95,Banerjee03}. La plupart sont basés sur une ressource lexicale, comme WordNet \cite{wordnet}. Une première famille d'approches se base sur la structure de cette ressource externe (souvent un réseau sémantique ou un graphe orienté) et la similarité est alors calculée en fonction des distances des chemins reliant les concepts dans cette structure \cite{Budanitsky06}. Cependant, comme nous l'avons déjà dit précédemment, la structure de ces ressources ne reflète pas forcement la sémantique des images, et ce type de mesures ne semble donc pas adapté à notre problématique. Une approche alternative pour mesurer le degré de similarité sémantique entre deux concepts est d'utiliser la définition textuelle associée à ces concepts. Dans le cas de WordNet, ces définitions sont connues sous le nom de glosses. Par exemple, Banerjee et Pedersen \cite{Banerjee03} ont proposé une mesure de proximité sémantique entre deux concepts qui est basée sur le nombre de mots communs (chevauchements) dans leurs définitions (glosses).

Dans notre approche, nous avons utilisé la mesure de similarité proposée par \cite{Patwardhan06}, qui se base sur WordNet et l'exploitation des vecteurs de co-occurrences du second ordre entre les glosses. Plus précisément, dans une première étape un espace de mots de taille $\mathcal{P}$ est construit en prenant l'ensemble des mots significatifs utilisés pour définir l'ensemble des synsets\footnote{Synonym set: composante atomique sur laquelle repose WordNet, composée d'un groupe de mots interchangeables dénotant un sens ou un usage particulier. A un concept correspond un ou plusieurs synsets.} de WordNet. Ensuite, chaque concept $c_i$ est représenté par un vecteur $\overrightarrow{w}_{c_i}$ de taille $\mathcal{P}$, où chaque \emph{ième} élément de ce vecteur représente le nombre d'occurrences du \emph{ième} mot de l'espace des mots dans la définition de $c_i$. La similarité sémantique entre deux concepts $c_i$ et $c_j$ est alors mesurée en utilisant la similarité cosinus entre $\overrightarrow{w}_{c_i}$ et $\overrightarrow{w}_{c_j}$:
\begin{equation}\label{Equ:GlossVector_sim}
    \eta(c_i,c_j)=\frac{\overrightarrow{w}_{c_i}\cdot\overrightarrow{w}_{c_j}}{|\overrightarrow{w}_{c_i}| |\overrightarrow{w}_{c_j}|}
\end{equation}

Certaines définitions de concepts dans WordNet sont très concises et rendent donc cette mesure peu fiable. En conséquence, les auteurs de \cite{Patwardhan06} ont proposé d'étendre les glosses des concepts avec les glosses des concepts situés dans leur voisinage d'ordre 1. Ainsi, pour chaque concept $c_i$ l'ensemble $\Psi_{c_i}$ est défini comme l'ensemble des glosses adjacents connectés au concept $c_i$ ($\Psi_{c_i}$=\{gloss($c_i$), gloss(hyponyms($c_i$)), gloss(meronyms($c_i$)), etc.\}). Ensuite pour chaque élément $x$ (gloss) de $\Psi_{c_i}$ , sa représentation $\overrightarrow{w}_{x}$ est construite comme expliqué ci-dessus. La mesure de similarité entre deux concepts $c_i$ et $c_j$ est alors définie comme la somme des cosinus individuels des vecteurs correspondants:
\begin{equation}\label{Equ:Gloss_sim}
\theta(c_i,c_j)=\frac{1}{|\Psi_{c_i}|}\sum_{x \in \Psi_{c_i}, y\in \Psi_{c_j}} \frac{\overrightarrow{w}_{x}\cdot\overrightarrow{w}_{y}}{|\overrightarrow{w}_{x}| |\overrightarrow{w}_{y}|}
\end{equation}
où $|\Psi|=|\Psi_i|=|\Psi_j|$.

Enfin, chaque concept dans WordNet peut correspondre à plusieurs sens (synsets) qui diffèrent les uns des autres dans leur position dans la hiérarchie et leur définition. Une étape de désambiguïsation est donc nécessaire pour l'identification du bon synset. Par exemple, la similarité entre "Souris" (animal) et "Clavier" (périphérique) diffère largement de celle entre "Souris" (périphérique) et "Clavier" (périphérique). Ainsi, nous calculons d'abord la similarité conceptuelle entre les différents sens (synset) de $c_i$ et $c_j$. La valeur maximale de similarité est ensuite utilisée pour identifier le sens le plus probable de  ces deux concepts, i.e. désambigüiser $c_i$ et $c_j$. La similarité conceptuelle est alors calculée par la formule suivante:
\begin{equation}\label{Words_sim}
\pi(c_i,c_j)=\underset{\delta_i \in s(c_i), \delta_j \in s(c_j)}{\operatorname{argmax}} \theta(\delta_i,\delta_j)
\end{equation}
où $s(c_x)$ est l'ensemble des synsets qu'il est possible d'associer aux différents sens du concept $c_x$.

\subsection{Similarité Contextuelle}

Comme cela a été expliqué dans la section \ref{sec:Disc}, l'information liée au contexte d'apparition des concepts est très importante dans un cadre d'annotation d'images. En effet, cette information, dite contextuelle, permet de relier des concepts qui apparaissent souvent ensemble dans des images ou des mêmes types d'images, bien que sémantiquement éloignés du point de vue taxonomique. De plus, cette information contextuelle peut aussi permettre d'inférer des connaissances de plus haut niveau sur l'image. Par exemple, si une photo contient "Mer" et "Sable", il est probable que la scène représentée sur cette photo est celle de la plage. Il semble donc important de pouvoir mesurer la similarité contextuelle entre deux concepts. Contrairement aux deux mesures de similarité précédentes, cette mesure de similarité contextuelle dépend du corpus, ou plus précisément dépend de la répartition des concepts dans le corpus.

Dans notre approche, nous modélisons la similarité contextuelle entre deux concepts $c_i$ et $c_j$ par l'information mutuelle PMI \cite{Church90} (Pointwise mutual information) $\rho(c_i,c_j)$:
\begin{equation}
\label{Equ:joint_pro}
    \rho(c_i,c_j)=  \log \frac{P(c_i,c_j)}{P(c_i)P(c_j)}
\end{equation}
où, $P(c_i)$ est la probabilité d'apparition de $c_i$, et $P(c_i,c_j)$ est la probabilité jointe de $c_i$ et de $c_j$. Ces probabilités sont estimées en calculant les fréquences d'occurrence et de cooccurrence des concepts $c_i$ et $c_j$ dans la base d'images.

Étant donné $\mathcal{N}$ le nombre total de concepts dans notre base d'images, $\mathcal{L}$ le nombre total d'images, $n_i$ le nombre d'images annotées par $c_i$ (fréquence d'occurrence de $c_i$) et $n_{ij}$ le nombre d'images co-annotées par $c_i$ et $c_j$, les probabilités précédentes peuvent être estimées par:
\begin{equation}\label{Pci}
\begin{array}{cc}
    \widehat{P(c_i)}=\frac{n_i}{\mathcal{L}}, & \widehat{P(c_i,c_j)}=\frac{n_{ij}}{\mathcal{L}}\\
\end{array}
\end{equation}

Ainsi:
\begin{equation}\label{Equ:ContextSim}
   \rho(c_i,c_j)= \log \frac{{\mathcal{L}*n_{ij}}}{n_i*n_j}
\end{equation}

$\rho(c_i,c_j)$ quantifie la quantité d'information partagée entre les deux concepts $c_i$ et $c_j$. Ainsi, si $c_i$ et $c_j$ sont des concepts indépendants, alors $P(c_i,c_j)=P(c_i)\cdot P(c_j)$ et donc $\rho(c_i,c_j)= log~1=0$. $\rho(c_i,c_j)$ peut être négative si $c_i$ et $c_j$ sont corrélés négativement. Sinon, $\rho(c_i,c_j) > 0$ et quantifie le degré de dépendance entre ces deux concepts. Dans ce travail, nous cherchons uniquement à mesurer la dépendance positive entre les concepts et donc nous ramenons les valeurs négatives de $\rho(c_i,c_j)$ à 0.

Enfin, afin de la normaliser dans l'intervalle [0,1], nous calculons la similarité contextuelle entre deux concepts $c_i$ et $c_j$ dans notre approche par:
\begin{equation}\label{Equ:ContextSim}
   \gamma(c_i,c_j)= \frac{\rho(c_i,c_j)}{-\log[\max(P(c_i),P(c_j))]}
\end{equation}

Il est à noter que la mesure PMI dépend de la distribution des concepts dans la base. Plus un concept est rare plus sa PMI est grande. Donc si la distribution des concepts dans la base n'est pas uniforme, il est préférable de calculer $\rho$ par:
\begin{equation}\label{Equ:rho}
    \rho(c_i,c_j)=  P(c_i,c_j)\log \frac{P(c_i,c_j)}{P(c_i)P(c_j)}
\end{equation}

\subsection{Mesure de Similarité Proposée}

Pour deux concepts donnés, les mesures de similarité visuelle, conceptuelle et contextuelle sont d'abord normalisées dans le même intervalle. La normalisation est faite par la normalisation Min-Max. Puis en combinant les mesures précédentes, nous obtenons la mesure de similarité sémantique adaptée à l'annotation suivante:
\begin{equation}
\label{equ:fusion}
    \phi(c_i,c_j)=\omega_1 \cdot \overline{\varphi}(c_i,c_j)+\omega_2 \cdot \overline{\pi}(c_i,c_j)+ \omega_3 \cdot \overline{\gamma}(c_i,c_j)
\end{equation}
où: $\sum_{i=1}^3 \omega_i =1$; $\overline{\varphi}(c_i,c_j)$, $\overline{\pi}(c_i,c_j)$ et $\overline{\gamma}(c_i,c_j)$ sont respectivement la similarité visuelle, la similarité conceptuelle et la similarité contextuelle normalisées.

Le choix des pondérations $\omega_i$ est très important. En effet, selon l'application ciblée, certains préféreront construire une hiérarchie spécifique à un domaine (qui représente le mieux une particularité d'un domaine ou d'un corpus), et pourront donc attribuer un plus fort poids à la similarité contextuelle ($\omega_3\nearrow$). D'autres pourront vouloir créer une hiérarchie générique, et devront donc donner plus de poids à la similarité conceptuelle ($\omega_2\nearrow$). Toutefois, si le but de la hiérarchie est plutôt de construire une plateforme pour la classification de concepts visuels, il est peut être avantageux de donner plus de poids à la similarité visuelle ($\omega_1\nearrow$).

\subsection{Règles pour la création de la hiérarchie}

La mesure proposée précédemment ne permet que de donner une information sur la similarité entre les concepts deux à deux. Notre objectif est de regrouper ces différents concepts dans une structure hiérarchique. Pour cela, nous définissons un ensemble de règles qui permettent d'inférer les relations d'hypernymie entre les concepts.

Nous définissons d'abord les fonctions suivantes sur lesquelles se basent nos règles de raisonnement:
\begin{itemize}
  \item $Closest(c_i)$ qui retourne le concept le plus proche de $c_i$ selon notre mesure:
\begin{equation}\label{Eq:Closest}
\begin{split}
Closest(c_i)= \underset{c_k \in \mathcal{C}\backslash\{c_i\}}{\operatorname{argmax}} \phi(c_i,c_k)
\end{split}
\end{equation}

\item $LCS(c_i,c_j)$ permet de trouver l'ancêtre commun le plus proche (\emph{Least Common Subsumer}) de $c_i$ et $c_j$ dans WordNet:
\begin{equation}\label{LCS}
\begin{split}
LCS(c_i,c_j)= \underset{c_l \in \{H(c_i)\cap H(c_j)\}}{\operatorname{argmin}} len(c_l,root)
\end{split}
\end{equation}
où $H(c_i)$ permet de trouver l'ensemble des hypernymes de $c_i$ dans la ressource WordNet, $root$ représente la racine de la hiérarchie WordNet et $len(c_x,root)$ renvoie la longueur du plus court chemin entre $c_x$ et $root$ dans WordNet.
\item $Hits_3(c_i)$ renvoie les 3 concepts les plus proche de $c_i$ au sens de la fonction $Closest(c_i)$.
\end{itemize}

\begin{figure}[!h]
\center
\subfigure[$1^{ere}$ Règle.]{
\includegraphics[width=7cm]{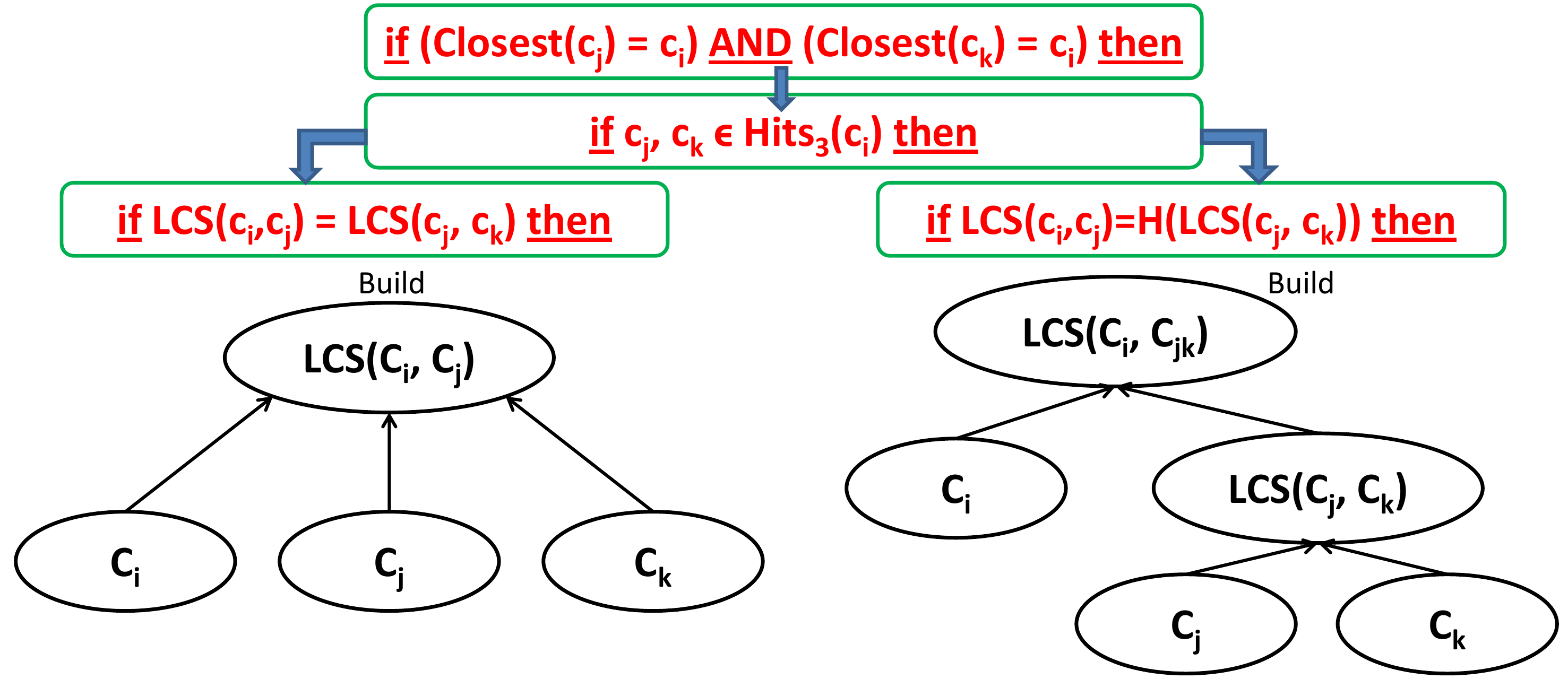}
\label{fig:rule1}}
\subfigure[$2^{ieme}$ Règle.]{
\includegraphics[width=4cm]{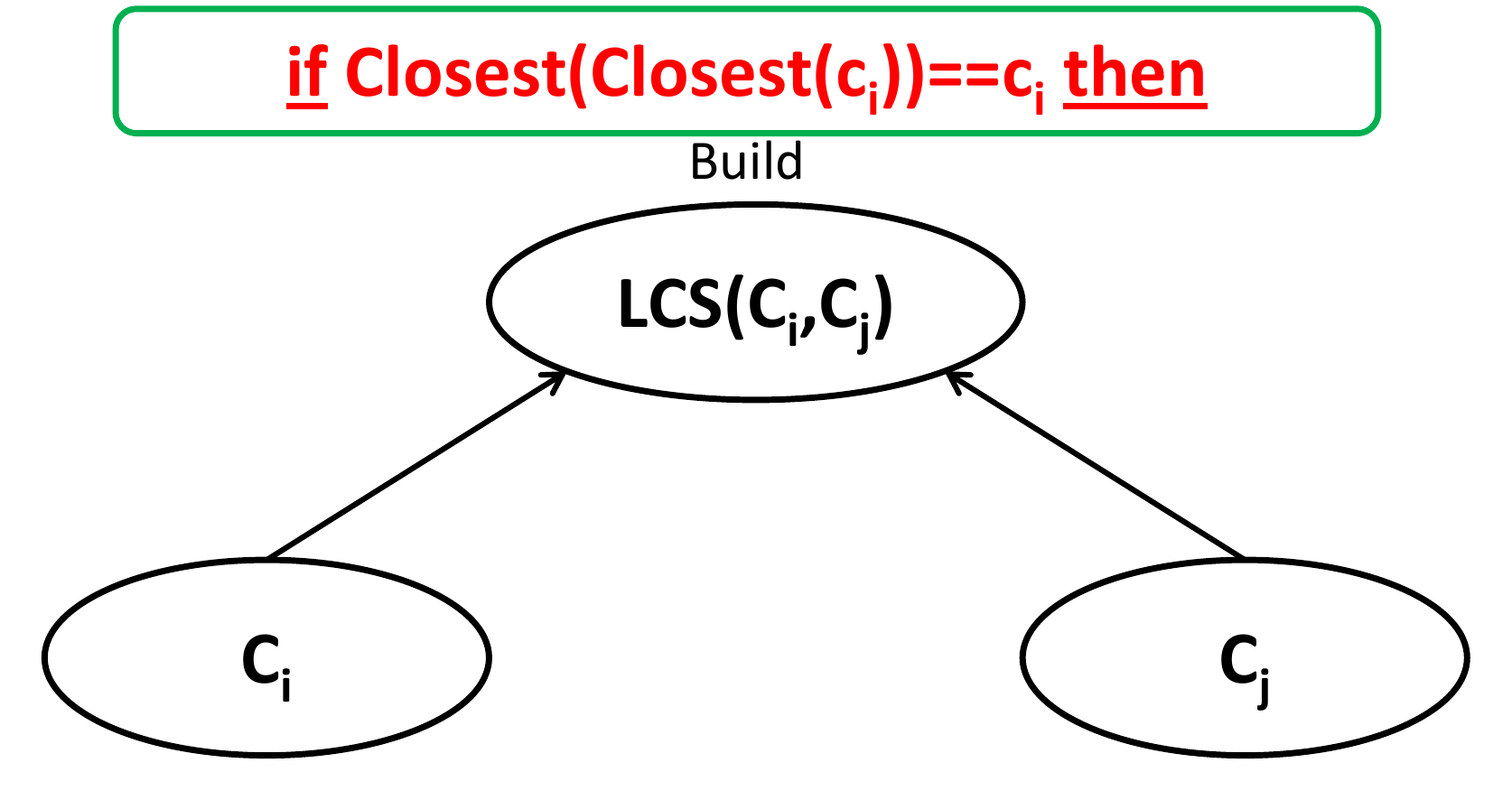}
\label{fig:rule2}}
\subfigure[$3^{ieme}$ Règle.]{
\includegraphics[width=7cm]{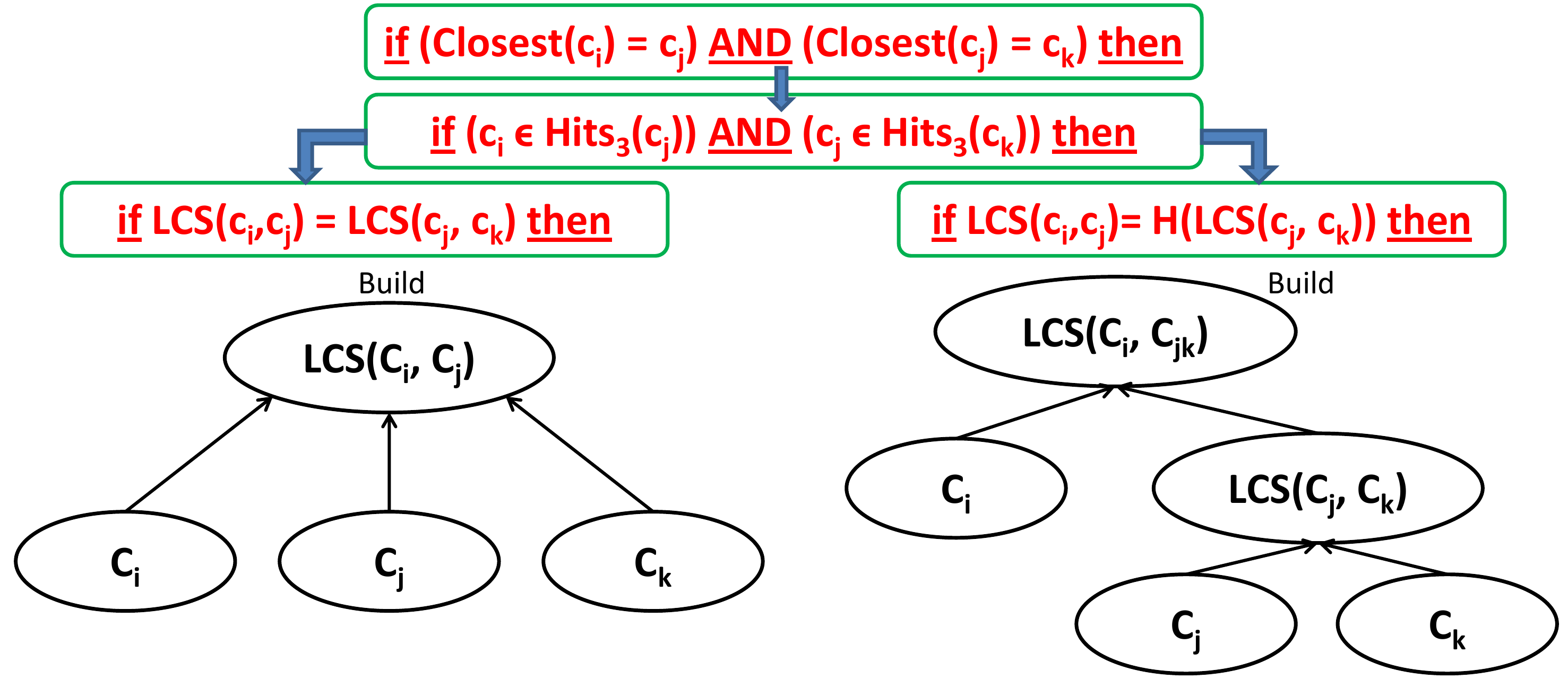}
\label{fig:rule3}}
\caption{Règles pour inférer les liens de parenté entre les différents concepts. En rouge les préconditions devant être satisfaites, en noir les actions de création de n\oe uds dans la hiérarchie.}
\label{fig:Regles}
\end{figure}

Nous définissons ensuite trois règles qui permettent d'inférer les liens de parenté entre les différents concepts. Ces différentes règles sont représentées graphiquement sur la figure \ref{fig:Regles}. Ces règles sont exécutées selon l'ordre décrit dans la figure \ref{fig:Regles}. La première règle vérifie si un concept $c_i$ est classé comme le plus proche par rapport à plusieurs concepts ($(Closest(c_j)=c_i), \forall j \in \{1,2,\cdots\}$). Si oui et si ces concepts $\{c_j\}, \forall j \in \{1,2,\cdots\}$, sont réciproquement dans $Hits_3(c_i)$, alors en fonction de leur LCS ils seront soit reliés directement à leur LCS ou dans une structure à 2 niveaux, comme illustré dans Figure \ref{fig:rule1}. Dans la seconde, si $(Closest(c_i)=c_j)$ et $(Closest(c_j)=c_i)$ (peut aussi être écrite $Closest(Closest(c_i))=c_i$) alors $c_i$ et $c_j$ sont fortement apparentés et seront reliés à leur LCS. La troisième règle concerne le cas où $(Closest(c_i)=c_j)$ et $(Closest(c_j)=c_k)$  - voir Figure \ref{fig:rule3}.

La construction de la hiérarchie suit une approche ascendante (i.e. commence à partir des concepts feuilles) et utilise un algorithme itératif jusqu'à atteindre le n\oe ud racine. Étant donné un ensemble de concepts associés aux images dans un ensemble d'apprentissage, notre méthode calcule la similarité $\phi(c_i,c_j)$ entre toutes les paires de concepts, puis relie les concepts les plus apparentés tout en respectant les règles définies précédemment. La construction de la hiérarchie se fait donc pas-à-pas en ajoutant un ensemble de concepts inférés des concepts du niveau inférieur. On itère le processus jusqu'à ce que tous les concepts soient liés à un n\oe ud racine.

\section{Résultats Expérimentaux}

Pour valider notre approche, nous comparons la performance d'une classification plate d'images avec une classification hiérarchique exploitant la hiérarchie construite avec notre approche sur les données de Pascal VOC'2010 (11 321 images, 20 concepts).

\subsection{Représentation Visuelle}
Pour calculer la similarité visuelle des concepts, nous avons utilisé dans notre approche le modèle de sac-de-mots visuels (Bag of Features) (BoF). Le modèle utilisé BoF est construit comme suit: détection de caractéristiques visuelles à l'aide des détecteurs DoG de Lowe \cite{Lowe99}, description de ces caractéristiques visuelles en utilisant le descripteur SIFT \cite{Lowe99}, puis génération du dictionnaire eu utilisant un K-Means. Le dictionnaire généré est un ensemble de caractéristiques supposées être représentatives de toutes les caractéristiques visuelles de la base. Étant donnée la collection de patches (point d'intérêt) détectés dans les images de l'ensemble d'apprentissage, nous générons un dictionnaire de taille $D=1000$ en utilisant l'algorithme k-Means. Ensuite, chaque patch dans une image est associé au mot visuel le plus similaire dans le dictionnaire en utilisant un arbre KD. Chaque image est alors représentée par un histogramme de $1000$ mots visuels (1000 étant la taille du codebook), où chaque bin dans l'histogramme correspond au nombre d'occurrences d'un mot visuel dans cette image.

\subsection{Pondération}

\begin{figure*}[!t]
\center
\includegraphics[width=18cm]{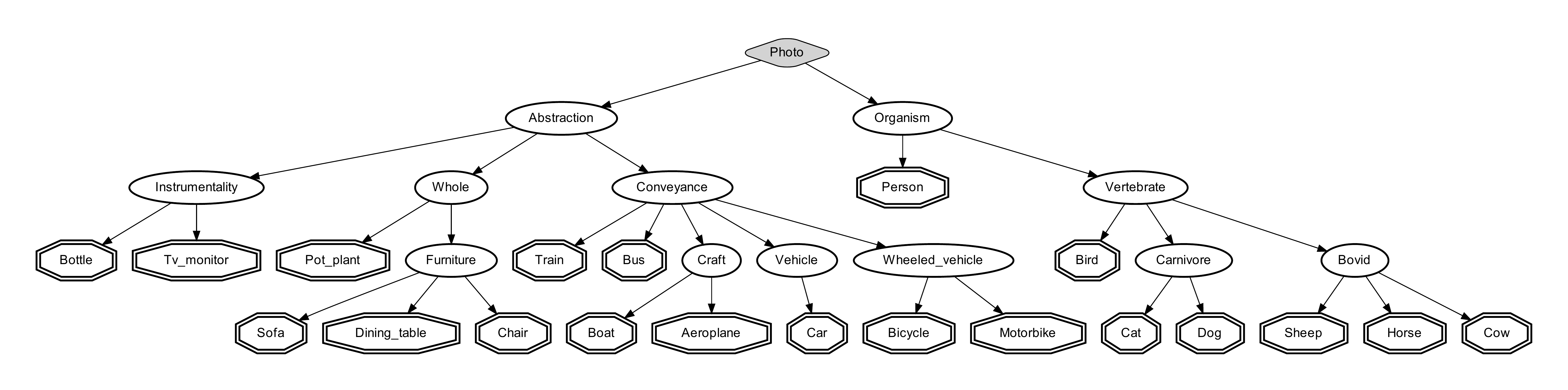}
\caption{La hiérarchie sémantique construite sur les données de Pascal VOC en utilisant la mesure proposée et les règles de construction. Les n\oe uds en double octogone sont les concepts de départ, le n\oe ud en diamant est la racine de la hiérarchie construite et les autres sont les n\oe uds inférés.\,\,\,\,\,\,\,\, $\phi(c_i,c_j)=0.4 \cdot \overline{\varphi}(c_i,c_j)+0.3 \cdot \overline{\pi}(c_i,c_j)+ 0.3 \cdot \overline{\gamma}(c_i,c_j)$} \label{fig:Taxonomy}
\end{figure*}

Comme ce travail vise à construire une hiérarchie adaptée à l'annotation et la classification d'images, nous avons fixé les facteurs de pondération de manière expérimentale comme suit : $\omega_1=0.4$, $\omega_2= 0.3$, et $\omega_3=0.3$. Nos expérimentations sur l'impact des poids ($\omega_i$) ont également montré que la similarité visuelle est plus représentative de la similarité sémantique des concepts, comme cela est illustré sur la figure \ref{fig:Taxonomy} avec la hiérarchie produite. Cette hiérarchie est construite sur les données de Pascal VOC'2010.

\subsection{Evaluation}

\begin{figure*}[!t]
\center
\includegraphics[width=12cm]{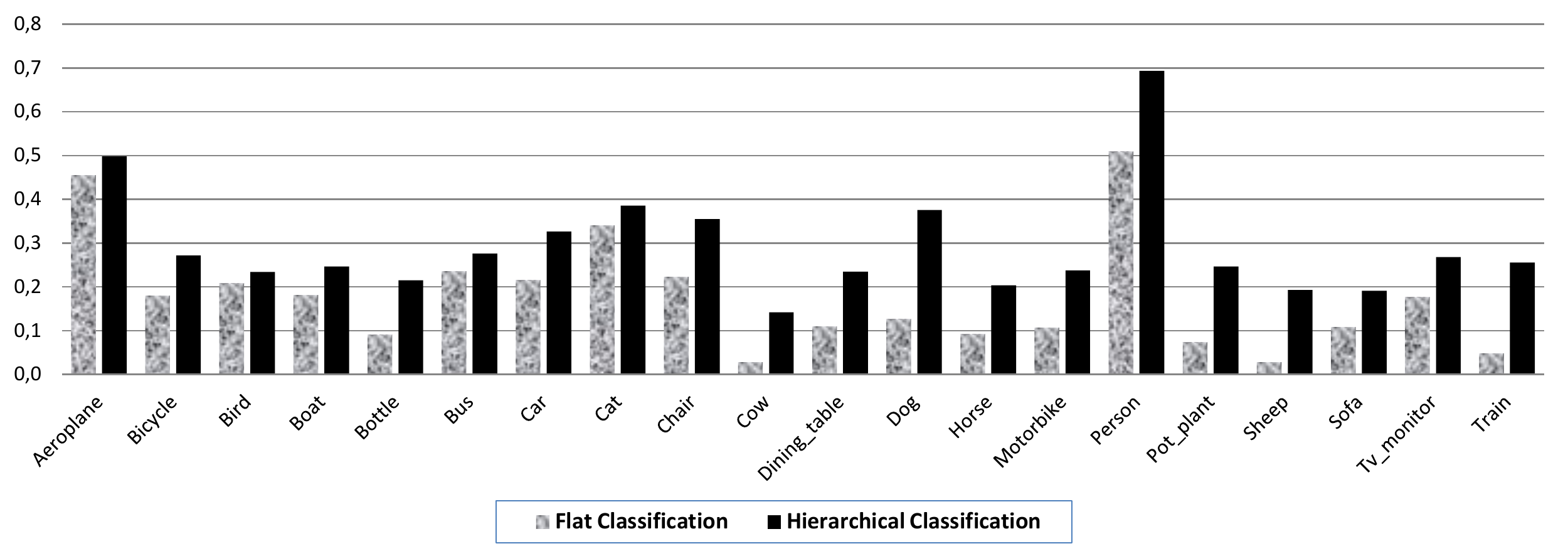}
\caption{Comparaison de la Précision Moyenne (AP) entre la classification plate et hiérarchique sur les données de Pascal VOC'2010.} \label{fig:comparaison}
\end{figure*}

\begin{figure}[!h]
\center
\subfigure[Concept Person.]{
\includegraphics[width=3.87cm]{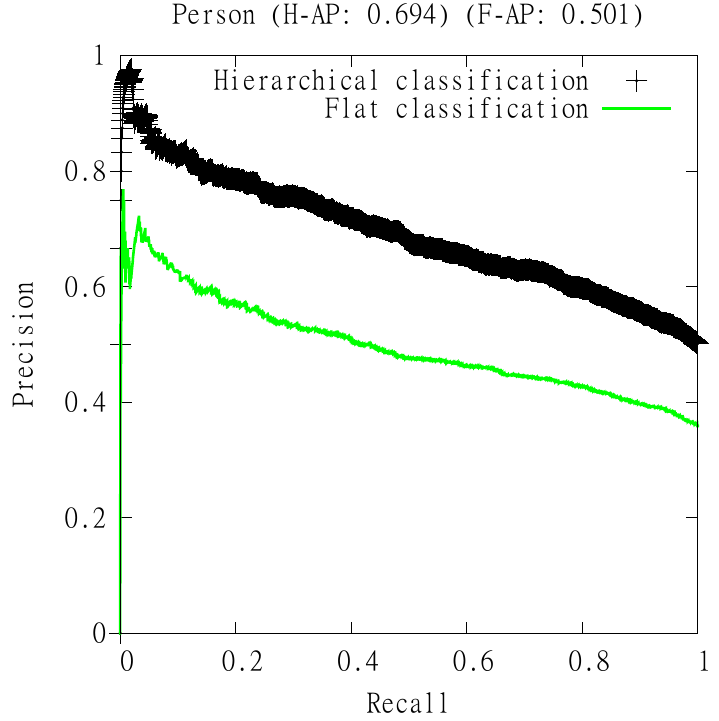}
\label{fig:subfig1}
}
\subfigure[Concept Tv\_monitor.]{
\includegraphics[width=3.87cm]{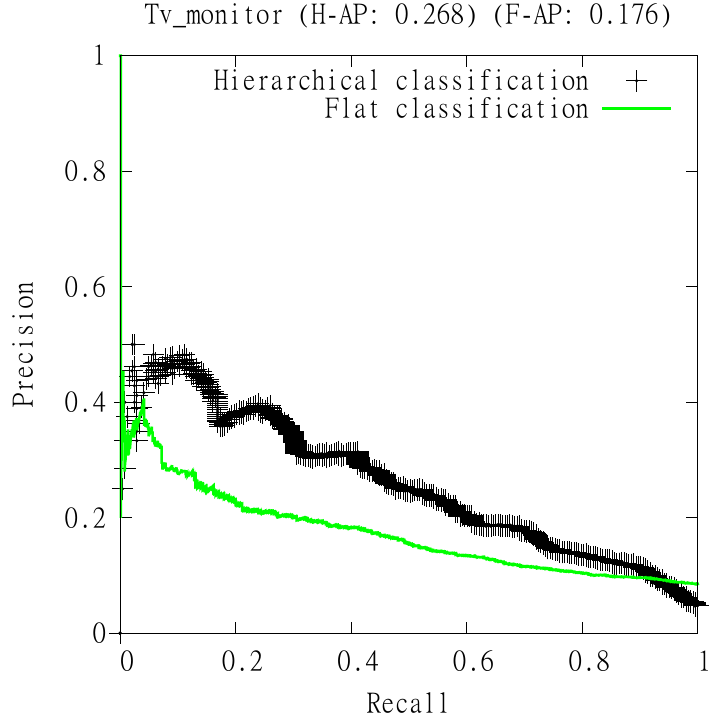}
\label{fig:subfig2}
}
\caption{Courbes Rappel/Précision pour la classification hiérarchique (en +) et plate (en trait) pour les concepts "Personne" et "TV\_Monitor".} \label{fig:Precision_recall}
\end{figure}

Pour évaluer notre approche, nous avons utilisé 50\% des images du challenge Pascal VOC'2010 pour l'apprentissage des classifieurs et les autres pour les tests. Chaque image peut appartenir à une ou plusieurs des 20 classes (concepts) existantes. La classification plate est faite par l'apprentissage de $\mathcal{N}$ SVM binaires un-contre-tous, où les entrées sont les représentations en BoF des images de la base et les sorties sont les réponses du SVM pour chaque image (1 ou -1) - pour plus de détails voir la section \ref{sec:Similarité_visuelle}. Un problème important dans les données de Pascal VOC est que les données ne sont pas équilibrées, i.e. plusieurs classes ne contiennent qu'une centaine d'images positives parmi les 11321 images de la base. Pour remédier à ce problème, nous avons utilisé la validation croisée d'ordre 5 en prenant à chaque fois autant d'images positives que négatives.

La classification hiérarchique est faite par l'apprentissage d'un ensemble de ($\mathcal{N}$+$\mathcal{M}$) classifieurs hiérarchiques conformes à la structure de la hiérarchie décrite dans la figure \ref{fig:Taxonomy}. $\mathcal{M}$ est le nombre de nouveaux concepts créés lors de la construction de la hiérarchie. Pour l'apprentissage de chacun des concepts de la hiérarchie, nous avons pris toutes les images des n\oe uds fils (d'un concept donné) comme positives et toutes les images des n\oe uds fils de son ancêtre immédiat comme négatives. Par exemple, pour apprendre un classifieur pour le concept "Carnivore", les images de "Dog" et "Cat" sont prises comme positives et les images de "Bird", "Sheep", "Horse" et "Cow" comme négatives. Ainsi chaque classifieur apprend à différencier une classe parmi d'autres dans la même catégorie. Durant la phase de test de la classification hiérarchique et pour une image donnée, on commence à partir du n\oe ud racine et on avance par niveau dans la hiérarchie en fonction des réponses des classifieurs des n\oe uds intermédiaires, jusqu'à atteindre un n\oe ud feuille. Notons qu'une image peut prendre plusieurs chemins dans la hiérarchie. Les résultats sont évalués avec les courbes rappel/précision et le score de précision moyenne.

La Figure \ref{fig:comparaison} compare les performances de nos classifieurs hiérarchiques avec les performances de la classification plate. L'utilisation de la hiérarchie proposée comme un cadre de classification hiérarchique assure des meilleures performances qu'une classification plate, avec une amélioration moyenne de +8.4\%. Notons que ces résultats sont obtenus en n'utilisant que la moitié des images du jeu d'apprentissage de Pascal VOC. En effet, en l'absence des images de test utilisées dans le challenge, nous avons utilisé le reste de l'ensemble d'apprentissage pour faire les tests. Nous avons aussi inclus les images marquées comme difficiles dans les évaluations de notre méthode. La précision moyenne de notre classification hiérarchique est de 28,2\%, alors que la classification plate reste à 19,8\%. On peut donc remarquer une nette amélioration des performances avec l'utilisation de la hiérarchie proposée. La Figure \ref{fig:Precision_recall} montre les courbes de rappel/précision des concepts "Personne" et "TV\_Monitor" en utilisant la classification hiérarchique et plate. Une simple comparaison entre ces courbes montre que la classification hiérarchique permet d'avoir un meilleur rendement à tous les niveaux de rappel. Cependant, il serait intéressant de tester notre approche sur une plus grande base, avec plus de concepts, pour voir si la hiérarchie construite pour la classification des images passe à l'échelle.

\section{Conclusion}

Ce papier présente une nouvelle approche pour construire automatiquement des hiérarchies adaptées à l'annotation sémantique d'images. Notre approche est basée sur une nouvelle mesure de similarité sémantique qui prend en compte la similarité visuelle, conceptuelle et contextuelle. Cette mesure permet d'estimer une
similarité sémantique entre concepts adaptée à la problématique de l'annotation. Un ensemble de règles est proposé pour ensuite effectivement relier les concepts entre eux selon la précédente mesure et leur ancêtre commun le plus proche dans WordNet. Ces concepts sont ensuite structurés en hiérarchie. Nos expériences ont montré que notre méthode fournit une bonne mesure pour estimer la similarité des concepts, qui peut aussi être utilisée pour la classification d'images et/ou pour raisonner sur le contenu d'images. Nos recherches futures porteront sur l'évaluation de notre approche sur des plus grandes bases d'images (MirFlicker et ImageNet) et sa comparaison avec l'état de l'art.

\bibliographystyle{abbrv}
\bibliography{biblio}

\begin{thebibliography}{10}

\bibitem{Banerjee03}
S.~Banerjee and T.~Pedersen.
\newblock Extended gloss overlaps as a measure of semantic relatedness.
\newblock In {\em {International Joint Conference on Artificial Intelligence
  (IJCAI'03)}}, 2003.

\bibitem{Bannour09CORIA}
H.~Bannour.
\newblock Une approche sémantique basée sur l'apprentissage pour la recherche
  d'image par contenu.
\newblock In {\em {COnférence en Recherche d'Infomations et Applications
  (CORIA'09)}}, pages 471--478, 2009.

\bibitem{Bannour11}
H.~Bannour and C.~Hudelot.
\newblock Towards ontologies for image interpretation and annotation.
\newblock In {\em {Content-Based Multimedia Indexing (CBMI'11)}}, pages 211
  --216, 2011.

\bibitem{BannourMMM12}
H.~Bannour and C.~Hudelot.
\newblock Building semantic hierarchies faithful to image semantics.
\newblock In {\em {advances in Multimedia Modeling (MMM'12)}}, volume 7131 of
  {\em Lecture Notes in Computer Science}, pages 4--15. Springer, 2012.

\bibitem{Barnard03}
K.~Barnard, P.~Duygulu, D.~Forsyth, N.~de~Freitas, D.~M. Blei, and M.~I.
  Jordan.
\newblock Matching words and pictures.
\newblock {\em Journal of Machine Learning Research}, 3:1107--1135, 2003.

\bibitem{Bart08}
E.~Bart, I.~Porteous, P.~Perona, and M.~Welling.
\newblock Unsupervised learning of visual taxonomies.
\newblock In {\em {Computer Vision and Pattern Recognition (CVPR'08)}}, 2008.

\bibitem{Blei04}
D.~M. Blei, T.~L. Griffiths, M.~I. Jordan, and J.~B. Tenenbaum.
\newblock Hierarchical topic models and the nested chinese restaurant process.
\newblock In {\em {Neural Information Processing Systems (NIPS'04)}}, 2004.

\bibitem{Budanitsky06}
A.~Budanitsky and G.~Hirst.
\newblock Evaluating wordnet-based measures of lexical semantic relatedness.
\newblock {\em {Computational Linguistics}}, 32:13--47, 2006.

\bibitem{Church90}
K.~W. Church and P.~Hanks.
\newblock Word association norms, mutual information, and lexicography.
\newblock {\em Comput. Linguist.}, 16:22--29, March 1990.

\bibitem{Vapnik95}
C.~Cortes and V.~Vapnik.
\newblock Support-vector networks.
\newblock {\em Machine Learning}, 20, 1995.

\bibitem{Deng10}
J.~Deng, A.~C. Berg, K.~Li, and L.~Fei-Fei.
\newblock What does classifying more than 10,000 image categories tell us?
\newblock In {\em European Conference on Computer Vision (ECCV'10)}, 2010.

\bibitem{Deng09}
J.~Deng, W.~Dong, R.~Socher, L.-J. Li, K.~Li, and L.~Fei-Fei.
\newblock Imagenet: A large-scale hierarchical image database.
\newblock In {\em {Computer Vision and Pattern Recognition (CVPR'09)}}, 2009.

\bibitem{Fan07}
J.~Fan, Y.~Gao, and H.~Luo.
\newblock Hierarchical classification for automatic image annotation.
\newblock In {\em {Conference on research and development in information
  retrieval (SIGIR'07)}}, pages 111--118, 2007.

\bibitem{Fan08IP}
J.~Fan, Y.~Gao, and H.~Luo.
\newblock Integrating concept ontology and multitask learning to achieve more
  effective classifier training for multilevel image annotation.
\newblock {\em {IEEE Transaction on Image Processing}}, 17(3), 2008.

\bibitem{Fan09}
J.~Fan, H.~Luo, Y.~Shen, and C.~Yang.
\newblock Integrating visual and semantic contexts for topic network generation
  and word sense disambiguation.
\newblock In {\em {ACM international Conference on Image and Video Retrieval
  (CIVR'09)}}, 2009.

\bibitem{wordnet}
C.~Fellbaum.
\newblock {\em {WordNet}: An Electronic Lexical Database}.
\newblock MIT Press, 1998.

\bibitem{Griffin08}
G.~Griffin and P.~Perona.
\newblock Learning and using taxonomies for fast visual categorization.
\newblock In {\em {Computer Vision and Pattern Recognition (CVPR'08)}}, 2008.

\bibitem{Hauptmann07}
A.~Hauptmann, R.~Yan, and W.-H. Lin.
\newblock How many high-level concepts will fill the semantic gap in news video
  retrieval?
\newblock In {\em {ACM international Conference on Image and Video Retrieval
  (CIVR'07)}}, pages 627--634, 2007.

\bibitem{Lavrenko03}
V.~Lavrenko, R.~Manmatha, and J.~Jeon.
\newblock A model for learning the semantics of pictures.
\newblock In {\em {Neural Information Processing Systems (NIPS'03)}}, 2003.

\bibitem{LI10}
L.-J. Li, C.~Wang, Y.~Lim, D.~M. Blei, and F.-F. Li.
\newblock Building and using a semantivisual image hierarchy.
\newblock In {\em {Computer Vision and Pattern Recognition (CVPR'10)}}, 2010.

\bibitem{Liu07}
Y.~Liu, D.~Zhang, G.~Lu, and W.-Y. Ma.
\newblock A survey of content-based image retrieval with high-level semantics.
\newblock {\em Pattern Recognition}, 40(1):262--282, 2007.

\bibitem{Lowe99}
D.~G. Lowe.
\newblock Object recognition from local scale-invariant features.
\newblock In {\em {International Conference on Computer Vision (ICCV'99)}},
  1999.

\bibitem{Marszalek07}
M.~Marszalek and C.~Schmid.
\newblock Semantic hierarchies for visual object recognition.
\newblock In {\em {Computer Vision and Pattern Recognition (CVPR'07)}}, pages
  1--7, 2007.

\bibitem{LSCOM06}
M.~Naphade, J.~R. Smith, J.~Tesic, S.-F. Chang, W.~Hsu, and L.~Kennedy.
\newblock Large-scale concept ontology for multimedia.
\newblock {\em {IEEE MultiMedia}}, 13:86--91, 2006.

\bibitem{Patwardhan06}
S.~Patwardhan and T.~Pedersen.
\newblock Using wordnet-based context vectors to estimate the semantic
  relatedness of concepts.
\newblock In {\em {Proceedings of the EACL 2006 Workshop on Making Sense of
  Sense: Bringing Computational Linguistics and Psycholinguistics Together}},
  pages 1--8, April 2006.

\bibitem{Resnik95}
P.~Resnik.
\newblock Using information content to evaluate semantic similarity in a
  taxonomy.
\newblock In {\em {International Joint Conferences on Artificial Intelligence
  (IJCAI'95)}}, 1995.

\bibitem{Bannour10}
L.~B. Romdhane, H.~Bannour, and B.~el~Ayeb.
\newblock Imiol: a system for indexing images by their semantic content based
  on possibilistic fuzzy clustering and adaptive resonance theory neural
  networks learning.
\newblock {\em {Applied Artificial Intelligence}}, 24(9):821--846, 2010.

\bibitem{Sivic08}
J.~Sivic, B.~C. Russell, A.~Zisserman, W.~T. Freeman, and A.~A. Efros.
\newblock Unsupervised discovery of visual object class hierarchies.
\newblock In {\em {Computer Vision and Pattern Recognition (CVPR'08)}}, 2008.

\bibitem{Smeulders00}
A.~W.~M. Smeulders, M.~Worring, S.~Santini, A.~Gupta, and R.~Jain.
\newblock Content-based image retrieval at the end of the early years.
\newblock {\em {IEEE Transaction Pattern Analysis and Machine Intelligence}},
  22:1349--1380, 2000.

\bibitem{Tousch11}
A.~Tousch, S.~Herbin, and J.-Y. Audibert.
\newblock Semantic hierarchies for image annotation: a survey.
\newblock {\em Pattern Recognition}, 2011.

\bibitem{Wei07}
X.-Y. Wei and C.-W. Ngo.
\newblock Ontology-enriched semantic space for video search.
\newblock In {\em {ACM Multimedia (MM'07)}}, pages 981--990, 2007.

\bibitem{Wu08}
L.~Wu, X.-S. Hua, N.~Yu, W.-Y. Ma, and S.~Li.
\newblock Flickr distance.
\newblock In {\em {ACM Multimedia (MM'08)}}, pages 31--40, 2008.

\bibitem{Yao09}
B.~Yao, X.~Yang, L.~Lin, M.~W. Lee, and S.~C. Zhu.
\newblock I2t: Image parsing to text description.
\newblock In {\em {Proceedings of IEEE}}, 2009.

\end{thebibliography}

\end{document}